\documentclass{ecai}
\usepackage{times}
\usepackage{graphicx}
\usepackage{latexsym}
\usepackage{float}
\ecaisubmission
\usepackage[utf8]{inputenc} 
\usepackage[T1]{fontenc}    
\usepackage{url}            
\usepackage{booktabs}       
\usepackage{amsfonts}       
\usepackage{nicefrac}       
\usepackage{microtype}      
\usepackage{amsmath}
\usepackage{amssymb}
\usepackage{subfigure}
\usepackage{multirow}
\usepackage{tabularx}
\usepackage{graphicx}
\usepackage{times}
\usepackage{graphicx,subfigure,amsmath,bbm,algorithm,algorithmic}
\usepackage[noend]{distribalgo}
\usepackage{xcolor}
\usepackage{caption}
\mathchardef\mhyphen="2D 
\DeclareMathOperator*{\argmin}{arg\,min} 
\title{Struct-MMSB: Mixed Membership Stochastic Blockmodels with Interpretable Structured Priors}

\author{Yue Zhang \and Arti Ramesh \institute{SUNY Binghamton,
USA, email: \{yzhan202, artir\}@binghamton.edu} }

\begin{document}

\maketitle

\begin{abstract}
The mixed membership stochastic blockmodel (MMSB) is a popular framework for community detection and network generation. It learns a low-rank mixed membership representation for each node across communities by exploiting the underlying graph structure. MMSB assumes that the membership distributions of the nodes are independently drawn from a Dirichlet distribution, which limits its capability to model highly correlated graph structures that exist in real-world networks. In this paper, we present a flexible richly structured MMSB model, \textit{Struct-MMSB}, that uses a recently developed statistical relational learning model, hinge-loss Markov random fields (HL-MRFs), as a structured prior to model complex dependencies among node attributes, multi-relational links, and their relationship with mixed-membership distributions. Our model is specified using a probabilistic programming templating language that uses weighted first-order logic rules, which enhances the model's interpretability. Further, our model is capable of learning latent characteristics in real-world networks via meaningful latent variables encoded as a complex combination of observed features and membership distributions. We present an expectation-maximization based inference algorithm that learns latent variables and parameters iteratively, a scalable stochastic variation of the inference algorithm, and a method to learn the weights of HL-MRF structured priors. We evaluate our model on six datasets across three different types of networks and corresponding modeling scenarios and demonstrate that our models are able to achieve an improvement of 15\% on average in test log-likelihood and faster convergence when compared to state-of-the-art network models.

\end{abstract}

\section{INTRODUCTION}
\label{sec:intro}

Modeling the complex and intricate interactions existing within a community is an important network science problem that has gained attention in the last decade. Perhaps one of the most commonly and widely used network generation and community detection model is mixed membership stochastic blockmodel (MMSB) \cite{airoldi2008mixed}, owing to its flexibility in modeling different kinds of networks and communities. MMSB models the node's membership in latent groups using a mixed-membership distribution, wherein the node can be part of multiple latent groups. This membership is used to generate the network structure. 
\noindent \paragraph{Related Work} 
MMSB has been adapted in various different ways to model complex network structures, such as hierarchical community structures \cite{ho2010infinite,sweet2014hierarchical,ho2011multiscale,zhang2017hierarchical}, dynamic networks \cite{fan2015dynamic,fu2009dynamic,xing2010state,ho2011evolving}, multi-relational graph structures \cite{yang2015multi,kim2012nonparametric,miller2009nonparametric,chen2016discriminative}, nodes participating in multiple communities \cite{zhou2015infinite,xu2016finding,xu2016finding,el2016scalable}, and feature-rich networks \cite{zhao2017leveraging,fan2017learning,yang2013community}. MMSB has also recently been extended to heterogenous networks \cite{huang2018}. Other probabilistic blockmodels have also been developed, such as infinite relational model (IRM) \cite{kemp2006learning}, Bi-LDA and Multi-LDA with extension to hierarchical Dirichlet processes (Multi-HDP) \cite{porteous2008multi},  binary space partitioning-tree process (BSP-tree) \cite{fan2018binary}, a generalized version of the Mondrian process \cite{roy2008mondrian}, and random function priors \cite{lloyd2012random}.

Recent advances in deep learning produced more powerful models for graph data, such as
graph neural networks (GNNs) \cite{ying2019gnn,ae482107de73461787258f805cf8f4ed,kipf2016semi}, GraphRNN \cite{conf/icml/YouYRHL18}, graph attention networks (GAT) \cite{velivckovic2017graph}. Bojchevski et al. developed NetGAN \cite{netgan}, which uses
the generative adversarial networks (GAN) framework to generate random walks on graphs and De Cao et al. \cite{de2018molgan} developed
MolGAN, which generates molecular graphs using the combination of a GAN framework and
a reinforcement learning objective. These models differ from MMSB and its variants in that they are not hand-engineered but can learn from data. But, the downside is the opaqueness and lack of interpretability of these models  \cite{ying2019gnn}, which Struct-MMSB seeks to address.

The above-mentioned blockmodels with the exception of a notable few ones such as the recently developed Copula-MMSB \cite{fan2016copula} assume that the membership indicator pairs are drawn independently; this limits the capability of the model to encode complex dependencies in the network structure. Even when the existing approaches relax this independence assumption, their applicability is restricted to only specific network structures (such as a-MMSB models assortative structure in networks) and are not solely versatile enough to handle the various kinds of dependencies present in real-world data, such as presence of additional features, multiple relationships, provision for learning meaningful latent variables. Further, as the models progress toward capturing complex dependencies, they are specified using complex functions that have limited interpretability. For example, the Copula-MMSB model uses the Copula function \cite{fan2016copula}, latent mixed-membership groups (LMMG)  uses the logistic function as priors \cite{kim2012latent}, and the BSP-tree uses a complex variant of the Mondrian process \cite{fan2018binary}. In addition to only capturing a specific kind of relationship, these priors are also harder to specify and understand for domain experts who may have a limited knowledge of machine learning. Incorporating correlations among the latent groups and memberships that can be easily interpreted provides the models with modeling power and possibility of application in domains that require careful specification of domain knowledge such as computational social science, bioinformatics, and other evolving application areas for societal good.

\begin{table*}[ht!]
    \begin{centering}
     {
     \footnotesize
         \caption{A comparison of general-purpose stochastic blockmodel frameworks}
    \begin{tabular}{p{5cm}p{2cm}p{4cm}p{1.5cm}p{2cm}}
    \toprule
Model &     {Dependencies} & Meaningful Latent Variables & Features & Scalability\\
    \midrule
    IRM \cite{kemp2006learning}& {$\times$} & {$\times$} & {$\times$} & {$\times$} \\  
    MMSB \cite{airoldi2008mixed} & {$\times$} & {$\times$} & {$\times$} & {\checkmark} \\ 
    LMMG \cite{kim2012latent}  & {$\times$} & {$\times$} & {\checkmark} & {\checkmark} \\ 
     a-MMSB \cite{gopalan2012scalable}  & {$\times$} & {$\times$} & {$\times$} & {\checkmark} \\ 
    Copula-MMSB \cite{fan2016copula} & {\checkmark} & {$\times$} & {$\times$} & {$\times$} \\ 
    \textbf{\textbf{Struct-MMSB}} (our approach) & {\checkmark} & {\checkmark} & {\checkmark} & {\checkmark} \\
    \bottomrule
    \end{tabular}
    \label{table:comparison}
    }
    \end{centering}
\end{table*}
\noindent \paragraph{Contributions} 
In this paper, we introduce a scalable and general purpose MMSB, \textit{Struct-MMSB}, by enhancing MMSB with a structured prior using a recently developed graphical model, hinge-loss Markov random fields (HL-MRFs).  Our approach, {Struct-MMSB}, is inspired from latent topic networks (LTN), a general-purpose latent Dirichlet allocation (LDA) using structured priors \cite{foulds2015latent}. Table \ref{table:comparison} gives a comparison of our model {Struct-MMSB} with other popular variants of MMSB. Our model possesses the capability to encode: 1) the relational dependencies among membership distributions, 2) additional node features, 3) multi-relational information, and their impact on the membership distributions using a probabilistic programming templating language, thus remaining interpretable and easy to specify custom network relationships. Further, our approach also incorporates the ability to encode meaningful latent variables, which can be learned as a complex combination of observed features, membership distributions, and group-group interaction probabilities, thus catering to many latent-variable modeling scenarios in computational social science problems. 

Next, we present algorithms for inference and learning in {Struct-MMSB}. We formulate an expectation maximization (EM) inference for inferring the expected value of latent variables, mixed-membership distributions of nodes in groups, and the group-group interaction probabilities. Then, we develop a scalable inference method using stochastic EM and show how to effectively incorporate relational dependencies, while remaining scalable to large networks. We then present a way to learn the weights of the first-order logical rules by maximizing the likelihood of the weights without additional ground truth data, thus allowing the model to learn the predictive ability of the logical rules in the data.

We demonstrate the versatility of our model in modeling different network modeling scenarios on data from six real-world networks and show that our model on average achieves a $15$\% better log-likelihood and a better prediction performance than the state-of-the-art MMSB variant, Copula-MMSB \cite{fan2016copula}, IRM \cite{kemp2006learning}, and MMSB \cite{airoldi2008mixed} and their multi-relational and online variants.

\section{BACKGROUND}
\label{sec:background}
In this section, we provide background on MMSB and HL-MRFs, and then proceed to show how to incorporate HL-MRFs as a structured prior in MMSB in {Struct-MMSB}. For ease of reference, we present a list of notations and their meanings to be used in the equations and derivations in this paper in Table \ref{table:results_90}.
\begin{table}
	\begin{centering}
	\normalsize{
		\caption{Notations for \textsc{Struct-MMSB}}
	\begin{tabular}{ |p{1.5cm}|p{6cm}|}
	\hline
	\footnotesize{N}&\footnotesize{number of nodes}\\
	\hline
		\footnotesize{p, q}&\footnotesize{specific nodes}\\
	\hline
\footnotesize{K}& \footnotesize{number of communities}\\
 	\hline
	\footnotesize{k, $k_1$, $k_2$}& \footnotesize{specific communities}\\
 	\hline
	\footnotesize{$\alpha$, $\beta^{(1)}$, $\beta^{(2)}$}& \footnotesize{hyperparameters of MMSB}\\
 	\hline
	\footnotesize{{$Y_{p, q}$}}& \footnotesize{link between nodes $p$ and $q$}\\
 	\hline
	\footnotesize{$z_{p \to q}$}& \footnotesize{membership indicator for sender node $p$ and receiver node $q$}\\	
 	\hline
	\footnotesize{{$\Pi$}}& \footnotesize{membership distributions of all the nodes}\\	
 	\hline
	\footnotesize{$\pi_p$}& \footnotesize{membership distribution of node $p$ across communities}\\	
 	\hline
	\footnotesize{$\pi^{(p)}_k$}& \footnotesize{value in the membership distribution of node $p$ for group $k$}\\	
 	\hline
	\footnotesize{$B$}& \footnotesize{matrix capturing interactions between communities}\\	
 	\hline
	\footnotesize{$B_{k_1, k_2}$}& \footnotesize{interaction between communities $k_1$ and $k_2$}\\	
	 \hline
	 	\footnotesize{$\Lambda$, $\lambda$}& \footnotesize{HL-MRF rule weights}\\	
	 \hline
	 	\footnotesize{$\psi_r$}& \footnotesize{{HL-MRF potential function for rule $r$}}\\	
	 \hline
		 	\footnotesize{$M(1)$}& \footnotesize{number of HL-MRF potentials for $\Pi$}\\	
	 \hline
		 	\footnotesize{$M(2)$}& \footnotesize{number of HL-MRF potentials for $B$}\\	
	\hline
		 	\footnotesize{$H^{(1)}$, $H^{(2)}$}& \footnotesize{latent variables in the HL-MRF prior}\\	
	\hline
		 	\footnotesize{$X$}& \footnotesize{observed features}\\	
	 \hline
	 	{\footnotesize{$\eta$}} & {\footnotesize{Lagrange coefficient}} \\
	 \hline
	\end{tabular}
   \label{table:results_90}
    }
	\end{centering}
\end{table}
\vspace{-0.3cm}
\subsection{Mixed Membership Stochastic Blockmodels (MMSB)}
\label{sec:structuredMMSB}
The generative process for MMSB is defined as follows.
\begin{itemize}
\item For each node $p \in N$:
\begin{itemize}
\item Draw a K dimensional mixed membership vector $\vec{\pi}_p$ ${\raise.17ex\hbox{$\scriptstyle\mathtt{\sim}$}}$ Dirichlet ($\vec{\alpha}$)
\end{itemize}
\item For each pair of nodes $(p, q)$ $\in$ $N \times N$
\begin{itemize}
\item Draw membership indicator for the initiator, $\vec{z}_{p\to q}$ ${\raise.17ex\hbox{$\scriptstyle\mathtt{\sim}$}}$ Multinomial ($\vec{\pi}_p$)
\item Draw membership indicator for the receiver, $\vec{z}_{q\to p}$ ${\raise.17ex\hbox{$\scriptstyle\mathtt{\sim}$}}$ Multinomial ( $\vec{\pi}_q$)
\item Sample the value of their interaction, $Y_{p, q}$ ${\raise.17ex\hbox{$\scriptstyle\mathtt{\sim}$}}$ Bernoulli($\vec{z}_{p\to q}^{\top}$$B$$\vec{z}_{q\to p}$), where $B$ ${\raise.17ex\hbox{$\scriptstyle\mathtt{\sim}$}}$ $Beta(\vec{\beta^{(1)}}, \vec{\beta^{(2)}})$
\end{itemize}
\end{itemize}
The independence assumptions implicit in the Dirichlet and Beta priors are what prevents MMSB from capturing complex dependencies. The priors are therefore our point of attack in developing a rich, flexible, and easy-to-encode stochastic blockmodel. In Struct-MMSB, we replace the flat Dirichlet and Beta priors with more expressive HL-MRF priors, we discuss HL-MRFs and their suitability as priors in MMSB below. 

%
%
\subsection{Hinge-loss Markov Random Fields}
\label{sec:hlmrf}
\vspace{-0.05cm}
HL-MRFs are a recently developed scalable class of continuous, conditional graphical models \cite{bach2017hinge}. HL-MRFs can be specified using \textit{Probabilistic Soft Logic (PSL)} \cite{bach2017hinge}, a first-order logic templating language. In PSL, random variables are represented as logical atoms and weighted rules define dependencies between them of the form:
 $\lambda : P(a) \land Q(a, b) \rightarrow R(b)$,
\normalsize
where \textit{P}, \textit{Q}, and \textit{R} are predicates, \textit{a} and \textit{b} are variables, and $\lambda$ is the weight associated with the rule. The weight of the rule $r$ indicates its importance in the HL-MRF model, which is defined as
\begin{align}
P(\mathit{Y}|\mathit{X}) &= \exp \Big (- \sum_{r=1}^M \lambda_r \psi_r(\mathit{Y}, \mathbf{X}) \Big )/Z(\lambda); \\
\hspace{0.4cm} Z(\lambda)& = \int_Y \exp \Big (- \sum_{r=1}^M \lambda_r \psi_r(\mathit{Y}, \mathit{X}) \Big ) \nonumber \\
 \psi_r(\mathit{Y}, \mathit{X}) &= \left( \max\{l_r(\mathbf{Y}, \mathit{X}),0 \}\right)^{\rho_r} 
\label{eqn:hl-potential}
\end{align}

where $\mathit{P(Y|X)}$ is the probability density function of a subset of logical atoms \textit{Y} given observed logical atoms \textit{X}, $\psi_r(\mathit{Y}, \mathit{X})$ is a \emph{hinge-loss potential} corresponding to an instantiation of a rule $r$, and is specified by a linear function $l_r$ and optional exponent $\rho_r \in \{1,2\}$, and $Z(\lambda)$ is the partition function. The logical conjunction of Boolean variables $X \land Y$ can be generalized to continuous variables using the hinge function $\textit{max}\{X+Y-1,0\}$, which is known as the Lukasiewicz t-norm. HL-MRFs admit tractable MAP inference regardless of the graph structure of the graphical model, making it feasible to reason over complex dependencies. This is possible because HL-MRFs operate on continuous random variables and encode dependencies using convex potential functions, so MAP inference in these models is a convex optimization problem. Giannini et al. \cite{giannini2018characterization} provide theoretical insight for understanding the convex characterization of the Lukasiewicz t-norms and Horn clauses in PSL \cite{bach2017hinge}. 

Our resulting model, Struct-MMSB, upon replacing the priors with HL-MRFs possess the following desirable qualities. They are structured and can capture complex relational dependencies among the nodes, their features, their membership distributions and group-group interactions using graphical model templates. These templates are simple weighted rules that capture the general characteristics of the network. The lucid nature of these rules make them inherently interpretable and intuitive and hence pave the way for ease of specification by domain experts. Further, these priors can be enriched using meaningful latent variables that further enhance their modeling power and intuitiveness. 

\section{{STRUCT-MMSB}}
\vspace{-0.25cm}
Our goal in designing Struct-MMSB is to create a lucid, easy-to-specify, and expressive generative model. We discuss the technical details of Struct-MMSB below.
\vspace{-0.2cm}
\subsection{Struct-MMSB graphical model}
\begin{figure}[!htb]
\centering
\includegraphics[width=0.5\columnwidth]{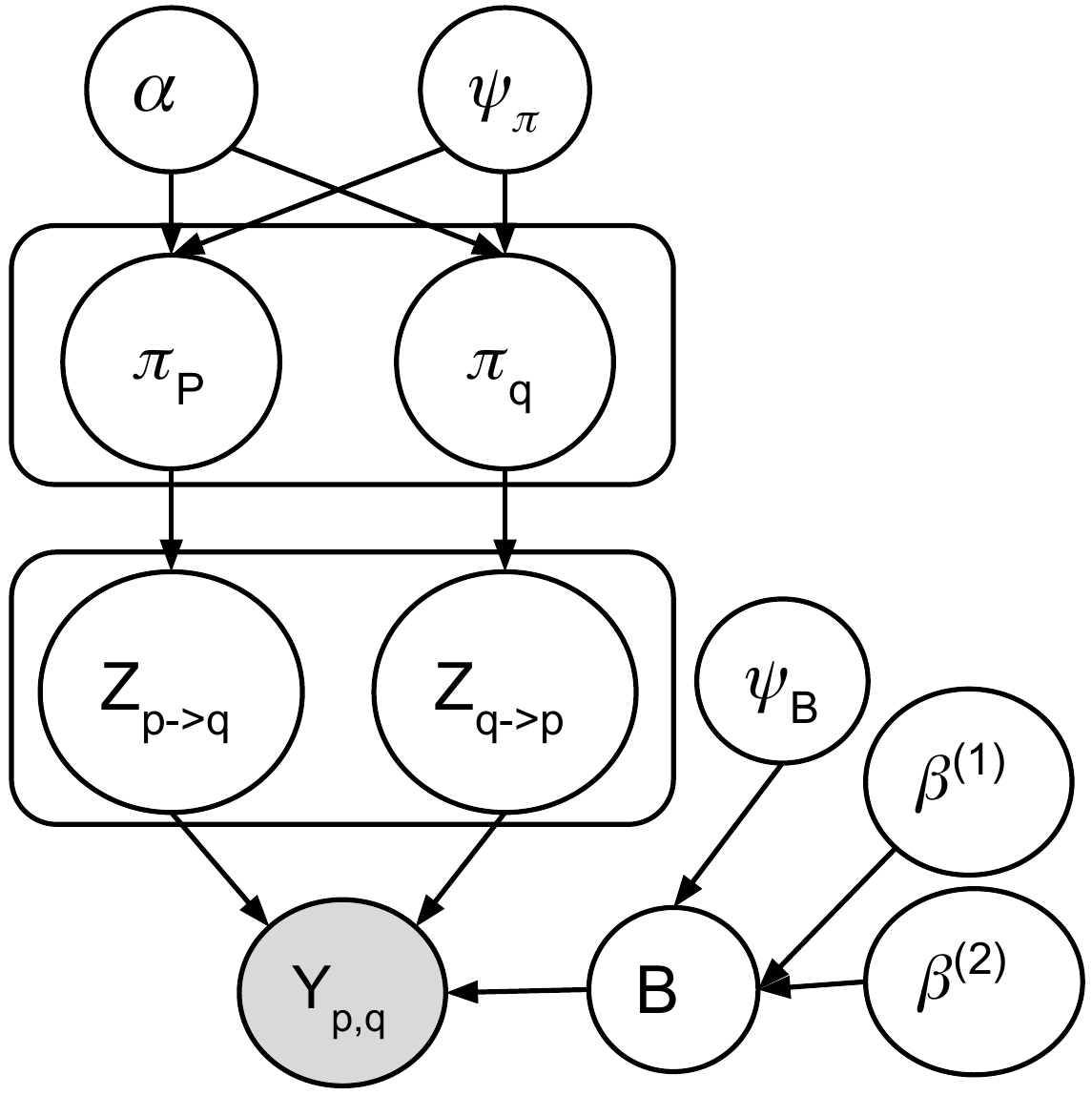}
\caption{Plate Diagram of {Struct-MMSB}}
\label{fig:plate}
\end{figure}

\vspace{-0.2cm}
To construct {Struct-MMSB}, we replace the Dirichlet priors and Beta priors in the MMSB generative process with HL-MRF potential function $\psi$. Figure \ref{fig:plate} shows the plate diagram of {Struct-MMSB}. The HL-MRF priors are indicated by $\psi_{\pi}$, which captures dependencies between the membership distribution $\pi$ of two nodes in the graph and $\psi_{B}$, which captures dependencies between groups in the group interaction matrix $B$. 
Before delving into the mathematical details, we provide some example dependencies that can be encoded using {Struct-MMSB}:\\
\noindent \textbf{Correlated nodes:} correlated($p$, $q$) $\land$ $\pi$($p$, $K$) $\rightarrow$ $\pi$($q$, $K$), if two nodes $p$ and $q$ are correlated, then their corresponding membership distributions ($\pi$) are correlated. \\
  \noindent \textbf{Latent characteristics modeled as latent variables:}  link($p$, $r$) $\land$ link($q$, $r$) $\rightarrow$ similar($p$, $q$), if two nodes $p$ and $q$ share common neighbor $r$, then they are similar.\\
  \noindent \textbf{Inter-group and Intra-group links:} correlated($p$, $q$) $\land$ $\pi$($p$, $K_1$) $\land$ B($K_1$, $K_2$) $\rightarrow$ $\pi$($q$, $K_2$), if two nodes $p$ and $q$ are correlated and the group interaction between $K_1$ and $K_2$ is high, then the value of the membership distribution $\pi$(p, $K_1$) is similar to $\pi$(q, $K_2$).


\subsection{Combining MMSB with HL-MRF priors}
In {Struct-MMSB}, the priors given by HL-MRFs define probability densities as follows: 
\begin{align*}
{\displaystyle
P(\Pi, H^{(1)} | X^{(1)}) \propto exp\bigg(-\sum_{j = 1}^{M^{(1)}} \lambda_j^{(1)} \psi_j^{(1)}(\Pi, X^{(1)}, H^{(1)})\bigg)} \\
{\displaystyle P(B, H^{(2)} | X^{(2)}) \propto exp\bigg(-\sum_{j = 1}^{M^{(2)}} \lambda_j^{(2)} \psi_j^{(2)}(B, X^{(2)}, H^{(2)})\bigg)}
\end{align*}
where $\psi_j^{(a)}$ are hinge-loss potentials, $M^{(a)}$ are number of hinge-loss potentials, and $X^{(a)}$ are observed features that are conditioned on, and $H^{(a)}$ are additional latent variables introduced by {Struct-MMSB}, and $\Pi$ and $B$ are MMSB parameters. 

The membership distributions \textbf{$\Pi$} are constrained to sum to $1$ for each node as in MMSB. For membership distributions that do not come from the HL-MRF priors, the default Dirichlet priors are used. We also add smoothed Dirichlet and Beta potentials for $\Pi$, $B$, and $1-B$, $\prod_{p,k}\pi_k^{p^{\alpha-1}}$, $\prod_{k_1,k_2} B_{k_1, k_2}^{\beta^{(1)} - 1}$, and $\prod_{k_1,k_2}(1- B_{k_1, k_2})^{\beta^{(2)} - 1}$, respectively. Incorporating the HL-MRF structured priors together with the smoothed Dirichlet and Beta potentials, the log-posterior of the variables of interest (parameters $\Pi$ and $B$, latent variables $H^{(1)}$ and $H^{(1)}$) is, 
\begin{flalign}
\label{eqn:logposterior}
& log P(\Pi, B, H^{(1)}, H^{(2)} | Y, \alpha, \beta^{(1)}, \beta^{(2)}, X^{(1)}, X^{(2)}, \Lambda) = \nonumber \\
& \hspace{0.3 cm} \sum_{p \in N, q \in N} log \bigg( \sum_{z_p \in K, z_q \in K}P(Y_{p,q}, z_p, z_q | \pi_p, \pi_q, B) \bigg) \nonumber \\
 & \hspace{0.3 cm} + \sum_{p \in N, k \in K} (\alpha - 1) log \pi_k^{p} \nonumber \\
 & \hspace{0.3 cm} - \sum_{j=1}^{M^{(1)}} \lambda_j^{(1)} \psi_j^{(1)}(\Pi, X^{(1)}, H^{(1)}) \nonumber \\
& \hspace{0.3 cm} + \sum_{k_1, k_2} \bigg( (\beta^{(1)} - 1) log B_{k_1, k_2} + (\beta^{(2)} - 1) log (1 - B_{k_1, k_2}) \bigg) \nonumber \\
 & \hspace{0.3 cm} - \sum_{j=1}^{M^{(2)}} \lambda_j^{(2)} \psi_j^{(2)}(B, X^{(2)}, H^{(2)}) + const.
\end{flalign}

\subsection{Training via EM}
\label{sec:training}

Our EM algorithm is motivated from  latent variable learning in HL-MRFs \cite{bach2015paired} and LTN \cite{foulds2015latent}. We train the model by maximum a posteriori (MAP) estimation, optimizing Eqn \ref{eqn:logposterior} with respect to $\Pi$, $B$, $H^{(1)}$, $H^{(2)}$. Since Eqn \ref{eqn:logposterior} cannot be optimized directly due to the sum inside the logarithm, we develop an EM algorithm that iteratively optimizes a lower bound arising from Jensen's inequality.  


Applying Jensen's inequality, we get the objective function to be, $ R(\Pi, B, X^{(1)}, X^{(2)}, H^{(1)}, H^{(2)}) \leq logP(\Pi, B, H | Y, \alpha, \beta^{(1)}, \beta^{(2)}, X, \Lambda) $, \normalsize where, 
\begin{flalign}
\label{eqn:objective}
& R(\Pi, B, X^{(1)}, X^{(2)}, H^{(1)}, H^{(2)}) = -\sum_{j=1}^{M^{(1)}} \lambda_j^{(1)} \psi_j^{(1)}(\Pi, X^{(1)}, H^{(1)}) \nonumber \\
& -\sum_{j=1}^{M^{(2)}} \lambda_j^{(2)} \psi_j^{(2)}(B, X^{(2)}, H^{(2)}) \nonumber \nonumber \\
&+ \sum_{p,k_1} \bigg( \sum_{q,k_2} (\gamma_{p,q,k_1,k_2} + \gamma_{q,p,k_2,k_1}) + \alpha - 1 \bigg) log \pi_{k_1}^{p} \nonumber \\
& + \sum_{k_1,k_2} \bigg(\sum_{p,q} \gamma_{p,q,k_1,k_2} Y_{p,q} + \beta^{(1)} - 1\bigg) log B_{k_1,k_2} \nonumber \\
& + \sum_{k_1,k_2} \bigg( \sum_{p,q} \gamma_{p,q,k_1,k_2} (1-Y_{p,q}) + \beta^{(2)}-1\bigg) log(1-B_{k_1,k_2}) \nonumber \\
&- \sum_{p,q,k_1,k_2} \gamma_{p,q,k_1,k_2} log \gamma_{p,q,k_1,k_2} + const
\end{flalign}
\normalsize
We define $\gamma_{p,q,k_1,k_2} \triangleq P(\vec{z}_{p \to q} = k_1, \vec{z}_{q \to p} = k_2 | \Pi^{(t)}, B^{(t)}, Y_{p,q})$, which encodes the distribution over the membership indicators $\vec{z}_{p \to q}$ and $\vec{z}_{q \to p}$ based on the previous parameter values of $\Pi$ and $B$. The algorithm consists of an E-step and an M-step, which are iterated until convergence. We first present the batch version of the algorithm, that iterates on all the data points for each E and M-step update and then present a stochastic variation that is scalable to large datasets.
\subsubsection{E-step}
To perform the E-step of the EM algorithm at iteration $t$, we first compute $\gamma_{p,q,k_1,k_2}$ in Equation \ref{eqn:objective}. 
\begin{flalign*}
 \gamma_{p,q,k_1,k_2} & \propto P(Y_{p,q} | z_p, z_q, \Pi^{(t)}, B^{(t)}) P(z_p, z_q| \Pi^{(t)}, B^{(t)}) \nonumber \\
 &= B_{k_1, k_2}^{Y_{p,q}} (1-B_{k_1, k_2})^{1-Y_{p,q}} \pi_{k_1}^{(p)} \pi_{k_2}^{(q)}
\end{flalign*}
Since the inference algorithm is iterative, the superscript (t) is used to refer to the previous/initialized values of $\Pi$ and B. 
\subsubsection{M-step}
We then perform the maximization step for $\Pi$ and $B$ parameters that are not involved in the HL-MRF prior, for which the update is identical to the M-step of the EM algorithm for standard MMSB. For $\Pi$, we add Lagrange terms $-\sum_{p} \eta_{k}^{\Pi} (\sum_{k} \pi_{k}^{p}-1) $ to constrain the parameter vectors to sum to one, where $\eta_{k}^{\Pi}$ is the Lagrange coefficient. To derive updates for $B$, we define $B$ and $1-B$ as $B^{(i)}, i \in \{0,1\}$, respectively, and add the Lagrange term $-\sum_{k_1k_2} \eta_{k_1k_2}^{B} (\sum_i B_{k_1k_2}^{(i)}-1)$ to constrain the $B_{k_1k_2}^{(i)}$ to sum to one, where $\eta_{k_1k_2}^{B}$ is the Lagrange coefficient. Taking derivatives and setting them to zero, we obtain the updates,
\begin{align*}
\pi_{k_1}^{p} &\propto \sum_{q,k_2} (\gamma_{p,q,k_1,k_2}+ \gamma_{q,p,k_2,k_1}) + \alpha -1; \nonumber \\
B^{(0)}_{k_1, k_2} &= \frac{ \sum_{p,q} \gamma_{p,q,k_1,k_2} Y_{p,q}+ \beta^{(1)}-1 }{\sum_{p,q} \gamma_{p,q,k_1,k_2}+ \beta^{(1)}+ \beta^{(2)}-2}
\end{align*}
\normalsize
Then, we optimize the lower bound jointly over the remaining $\Pi$ and $B$ parameters by fixing the parameters above that are not updated using HL-MRF priors. The hinge-loss potentials for $\Pi$ and $B$ can be optimized separately as evident from Eqn \ref{eqn:objective}. 
We minimize the negative of each of these two subproblems $-R(\Pi,H^{(1)},X^{(1)})$, $-R(B,H^{(2)},X^{(2)})$ using a consensus-optimization algorithm based on the alternating direction method of multipliers (ADMM). 

To accomplish this, we extend the consensus optimization algorithm by Bach et al. \cite{bach2017hinge} for MAP inference in HL-MRFs. The algorithm creates a local copy for each of the variables, thus dividing the original problem into independent subproblems that may be solved in parallel. The algorithm iteratively solves the independent subproblems and then updates the global consensus variables to be the average of the local copies. This is guaranteed to find the global optimum solution of the objective function. For more details, we refer the reader to \cite{bach2017hinge} and \cite{boyd2011distributed}. 

We extend this algorithm to include the objective function terms in the lower bound $R$ in Eqn \ref{eqn:objective} for $\Pi$ and $B$. For $\Pi$, the terms 1 and 3 in Eqn \ref{eqn:objective} are included and for $B$, the terms 2, 4, and 5 in Eqn \ref{eqn:objective} are included. The entropy term (term 6, $- \sum_{p,q,k_1,k_2} \gamma_{p,q,k_1,k_2} log \gamma_{p,q,k_1,k_2}$) and the constant term (term 7) are not included in the M-step objective as they do not have $\Pi$ or $B$ in them. To cast this as a consensus optimization problem, we create local copies of our variables of interest ($\Pi$ and $B$, denoted by $C_i$ in the ADMM update equation below), which we refer to as $c_i$, where $i$ refers to each $\pi_k^{p}$ and $B_{k_1, k_2}$, in the respective consensus optimization equations. The consensus optimization problem entails solving the subproblems with local copies $c_i$ independently for both the variables and then iterating till a consensus is reached, which is enforced via an equality constraint in the Lagrange term with coefficient $\eta_i$. For convenience, we refer to our objective terms as $A_i log c_i$, where $A_i$ denotes the coefficient in term 3 in Equation \ref{eqn:objective}, which equals to $(\sum_{q,k_2}(\gamma_{p,q,k_1,k_2}+\gamma_{q,p,k_2,k_1})+\alpha-1)$, and $c_i$ denotes the local copy of the variable.

The consensus ADMM update \cite{boyd2011distributed} for $c_i$ is given by,
\begin{align}
c_i = \argmin_{c_i^{\prime}}( -A_i log c_i^{\prime}+ \eta_i (c_i^{\prime} - C_i)+ \frac{\rho}{2} (c_i^{\prime}- C_i)^2) \nonumber
\end{align}
\normalsize
where $\rho$ is an ADMM step-size parameter, $c_i$ is the local copy of the variable used by the ADMM consensus optimization algorithm, and $C_i$ refers to the original variable.
In order to solve the ADMM update equation, we take its derivative with respect to $c_i$ and set it equal to zero,  $\frac{\partial J}{\partial c_i^{\prime}} = \rho c_i^{\prime 2}+ c_i^{\prime} (\eta_i- \rho C_i)- A_i =0$.



Solving this quadratic equation, we get two solutions. One of them is negative and can be discarded; $c_i$ is set to the positive solution. The Lagrange coefficients are updated using ADMM updates for Lagrangian dual, $\eta_i \leftarrow \eta_i+ \rho (c_i^{\prime}- C_i)$.





\vspace{-0.2cm}
\subsection{Stochastic EM}
\label{sec:stochastic}
To scale the batch model to large networks, we present a stochastic EM inference method. Our algorithm is motivated from the online EM algorithm \cite{liang2009online} and the stochastic optimization for \textit{a-}MMSB \cite{gopalan2012scalable}. We compute the expected sufficient statistics of $\Pi$ and $B$ and update it using randomly sampled node-pair $(p,q)$ from distribution $g(p,q)$ instead of the entire sample, by computing the stochastic natural gradient in the space of sufficient statistics. 



To obtain the stochastic EM algorithm, we first subsample the graph and compute the expected sufficient statistics for the sample, given the current settings of the $\Pi$ and $B$. We then update $\Pi$ and $B$ using these sufficient statistics. We introduce three global variables: $\theta^p_k$, $\phi_{k_1k_2}$, and $\phi'_{k_1k_2}$ to derive the stochastic EM updates. The membership distribution $\pi^p_k$ is equal to the normalized $\theta^p_k$. The affinity blockmodel $B^{(0)}_{k_1k_2}$ (which refers to $B$) is given by $\phi_{k_1k_2} / (\phi_{k_1k_2}+ \phi'_{k_1k_2})$, and $B^{(1)}_{k_1k_2}$ (which refers to $1-B$) equals $\phi'_{k_1k_2} / (\phi_{k_1k_2}+ \phi'_{k_1k_2})$. 


To derive the expected sufficient statistics of $\pi^p_k$ for a specific node $p$ and community $k_1$, we have,
\begin{flalign*}
& \bigg( \sum_{q,k_2} (\gamma_{p,q,k_1,k_2} + \gamma_{q,p,k_2,k_1})+ \alpha-1 \bigg) log \pi^p_{k_1} \\
& = \bigg( \sum_{q,k_2} g(p,q) \frac{1}{g(p,q)} (\gamma_{p,q,k_1,k_2} + \gamma_{q,p,k_2,k_1}) + \alpha-1 \bigg) log \pi^p_{k_1} \\
 & = \bigg(E_g [\frac{1}{g(p,q)} (\gamma_{p,q,k_1,k_2} + \gamma_{q,p,k_2,k_1})]+ \alpha-1 \bigg) log \pi^p_{k_1} 
\end{flalign*}
\normalsize
The intuition behind the above expected sufficient statistics comes from online EM for unsupervised models \cite{liang2009online} and stochastic variational EM for a-MMSB \cite{gopalan2012scalable}. Here, we sample a single node pair (p, q) and compute the $\gamma$ if our entire graph consisted of p and q, repeated $1/g(p,q)$ times. In practice, instead of a single node a mini-batch is sampled. Liang et al. \cite{liang2009online}  specify that updating on multiple data instances using a mini-batch adds more stability than updating after each sample. This translates to all $q$ in the mini-batch for a specific node of interest $p$. 


Hence, under each mini-batch, without HL-MRF priors, the expected sufficient statistics of $\theta^p_{k_1}$, $s_{p,k_1} = \frac{1}{g} \sum_{q,k2} (\gamma_{p,q,k_1,k_2} + \gamma_{q,p,k_2,k_1})+\alpha-1$. With HL-MRF priors, we first use ADMM to find local optimal value $\pi^{*p}_{k_1}$ under the current mini-batch, then the expected sufficient statistics is calculated as $s^{PSL}_{p,k_1} = \pi^{*p}_{k_1} \sum_{k_1} s_{p,k_1}$.
Similarly, we calculate the expected sufficient statistics for B. As $B_{k_1,k_2}$ and $1-B_{k_1,k_2}$ factorize into separate terms in Eqn \ref{eqn:objective}, we treat them as two different variables and calculate the expected sufficient statistics separately. For $B_{k_1, k_2}$, we have,
\begin{flalign*}
& \sum_{p,q} \gamma_{p,q,k_1,k_2} Y_{p,q} log B_{k_1,k_2} \\
&  =  \sum_{p,q} g(p,q) \bigg( \frac{1}{g(p,q)} \gamma_{p,q,k_1,k_2} Y_{p,q} \bigg)log B_{k_1,k_2} \\
& = E_g[ \frac{1}{g(p,q)} \gamma_{p,q,k_1,k_2} Y_{p,q}] log B_{k_1,k_2} 
\end{flalign*}
\normalsize
Under mini-batch, without HL-MRF priors, expected sufficient statistics of $\phi_{k_1,k_2}$ equals to, $s_{k_1k_2} = \frac{1}{g} \sum_{p,q} \gamma_{pqk_1k_2} Y_{p,q}+ \beta^{(1)}-1 $.
 Similarly, for $B'_{k_1, k_2} = 1-B_{k_1, k_2}$, we have,
\begin{flalign*}
& 
\sum_{p,q} \gamma_{p,q,k_1,k_2} (1-Y_{p,q}) log(B'_{k_1,k_2}) \\
& =  \sum_{p,q} g(p,q) \bigg( \frac{1}{g(p,q)} \gamma_{p,q,k_1,k_2} (1-Y_{p,q}) \bigg) log (B'_{k_1,k_2}) \\
&= E_g[ \frac{1}{g(p,q)} \gamma_{p,q,k_1,k_2} (1-Y_{p,q})] log (B'_{k_1,k_2})
\end{flalign*}
\normalsize
Under mini-batch, without HL-MRF priors, expected sufficient statistics of $\phi'_{k_1,k_2}$ equals to $s'_{k_1k_2} = \frac{1}{g} \sum_{p,q} \gamma_{p,q,k_1,k_2} (1-Y_{p,q})+\beta^{(2)}-1$. With HL-MRF priors, to compute the expected sufficient statistics for $\phi_{k_1k_2}$ and $\phi'_{k_1k_2}$, first we need to use ADMM to find local optimal value $B^{*}_{k_1k_2}$ under current mini-batch. Then, the  expected sufficient statistics for $\phi_{k_1k_2}$ becomes $s^{PSL}_{k_1k_2} = B^{*}_{k_1k_2} (s_{k_1k_2}+ s'_{k_1k_2})$ and for $\phi'_{k_1k_2}$, $s'^{PSL}_{k_1k_2} = (1-B^*_{k_1k_2}) (s_{k_1k_2}+ s'_{k_1k_2})$. Intuitively speaking, the expected sufficient statistics with HL-MRF priors re-assign the total mass without HL-MRF priors, $\sum s$, and make it proportional to the optimal value using the estimated $\pi^{*p}_{k_1}$ and $B^{*}_{k_1k_2}$ values from ADMM.

In our experiments, we use mini-batch instead of just one node pair. Evaluating the rewritten above equations for a node pair sampled from the graph gives a noisy but unbiased estimate of the batch model. We scale the expected sufficient statistics estimated from the subsample by $\frac{1}{g(p,q)}$ so that they are unbiased estimates of the true sufficient statistics as illustrated by Gopalan et al. \cite{gopalan2012scalable}. For instance, if we sample nodes uniformly at random, then $g(p,q) = \text{miniBatch}/N^2$. 

We calculate this as the difference between expected sufficient statistics  and previous value as follows
\begin{flalign}
\partial \theta^{t}_{pk}  & = s^{PSL}_{pk}- \theta^{t-1}_{pk} \nonumber\\
\partial \phi^{t}_{k_1k_2} & = s^{PSL}_{k_1k_2}- \phi^{t-1}_{k_1k_2} \nonumber\\
\partial \phi'^{t}_{k_1k_2} & = s'^{PSL}_{k_1k_2}- \phi'^{t-1}_{k_1k_2}
\end{flalign}
In the next step, we calculate the stochastic natural gradients computed from a sampled node pair $(p,q)$ or mini-batch and update the global variables, 
\vspace{-0.2cm}
\begin{flalign}
\theta^p_k & \leftarrow \theta^p_k+ \rho_t *{\partial \theta^{t}_{pk}}  \nonumber \\
\phi_{k_1k_2} & \leftarrow \phi_{k_1k_2}+ \rho_t *{\partial \phi^{t}_{k_1k_2}} \nonumber \\
\phi'_{k_1k_2} & \leftarrow \phi'_{k_1k_2}+ \rho_t *{\partial \phi'^{t}_{k_1k_2}}
\end{flalign}
\normalsize
where $\rho$ is global step size. Results from the stochastic approximation literature requires that $\sum_t \rho^2_t < \infty$ and $\sum_t \rho_t = \infty$ to guarantee convergence to a local optimum. We set $\rho_t \triangleq (\tau_0+t)^{-\kappa}$, where $\kappa \in (0.5,1]$ is the learning rate and $\tau_0 \geq 0$ downweights early iterations.

\paragraph{Distributed Implementation}
We implement a distributed computing environment using Apache Thrift framework for the stochastic EM inference. This further helps in scaling our models to larger datasets. We perform stochastic EM inference at each client side using subsampled mini-batch and then update the global parameters at server side.


\subsection{Weight learning}
The weights of the HL-MRF first-order-logic rules, $\Lambda$, can be selected based on domain knowledge. The challenge to learning weights arises because often there is no ground truth data for our variables of interest, $\Pi$ and $B$. Following Bach et al. \cite{bach2017hinge}, we present a maximum-likelihood estimation by maximizing the log-likelihood of the training data given parameter values to learn the weights. 
\begin{flalign}
\frac{\partial R}{\partial \Lambda_q}  = \frac{\partial log P(\Pi,H |X)}{\partial \Lambda_q} &= E_{\Lambda}[\Psi_q(X, H, \Pi)] - \Psi_q(X,H,\Pi) \nonumber\\
E_{\Lambda}[\Psi_q(X, H, \Pi)] &= \int_{\Pi} \Psi_q(X, H,\Pi) P(\Pi|X,H) \nonumber
\end{flalign}
\normalsize
We approximate the expectation using the value of the potential functions at the most probable setting of $\Pi$ with the current parameters. During weight learning, we treat latent variables $H$ as observations using the inferred value from the EM step. 


%
%

\section{EXPERIMENTS}
\label{sec:experiments}

We conduct experiments to answer the questions:\\

\noindent 1) {\it How our models perform on different network modeling scenarios?} To demonstrate the versatility of Struct-MMSB and its ability to model different real-world scenarios, we evaluate the performance of our model on a total of $6$ datasets and three network modeling scenarios: a) presence of additional features (\textit{case 1}), b) presence of multiple links (\textit{case 2}), and c) neighborhood similarity (\textit{case 3}),  \cite{leskovec2016snap}. We induce three different structured priors that correspond to the characteristics of each of these modeling scenarios. We compare the training and test log-likelihood and area under the ROC curve (AUC-ROC) of our models at convergence with MMSB  \cite{airoldi2008mixed} or its online/stochastic variant, IRM \cite{kemp2006learning}, and the state-of-the-art variant of MMSB (Copula-MMSB \cite{fan2016copula}), which has been compared to some other variants of MMSB and network models (MMSB, IRM, LFRM \cite{miller2009nonparametric},  and iMMM \cite{koutsourelakis2008finding}) and shown to be better. Since Copula-MMSB in the present form is not capable of handling multi-relational data, we extend Copula-MMSB to handle multi-relational data wherever appropriate; we refer to this model as Multi-Copula. We show that our model achieves better performance and faster convergence both under batch and stochastic inference when compared to the above-mentioned models.\\

\noindent 2) {\it How informative are the latent variables in our intuitive HL-MRF priors?} While performance is one important aspect, the more interesting contribution of our approach is the interpretable and intuitive nature of Struct-MMSB models, succinctly capturing domain knowledge in the different modeling scenarios. This interpretable nature is further enhanced by the presence of meaningful latent variables that can abstract the relationship between the features, nodes, and network connections. We show how to encode meaningful latent variables in all the three modeling scenarios: presence of additional features, presence of multiple links, and neighborhood similarity. We present qualitative analysis of the values learned by our latent variables and show that our models are able to encode and learn meaningful latent variable values that enhance the modeling power and interpretability of our model. 

\subsection{Experimental settings}
In all our experiments, our models use same initializations as standard MMSB. In the stochastic inference setting, our models also use the same mini-batches as online MMSB, so that they are exactly comparable. We initialize the hyper-parameters $\beta^{(1)}$ and $\beta^{(2)}$ of parameter $B$ by setting $\beta^{(1)}/ (\beta^{(1)}+ \beta^{(2)}) =$\textit{link-rate}, where \textit{link-rate}$ =link /(link + non\mhyphen link)$. We set  number of communities to $5$ for small datasets (feature-based similarity (case 1) and multi-relational experiments (case 2)), and to $8$ for the large datasets in link-based similarity (case 3) experiment. And, we randomly initialize the $\Pi$ and $B$ values. Statistically significant values with a rejection threshold of p = 0.05 are typed in \textbf{bold}.

\subsection{Case 1: feature-based similarity}
Here, we design models that take additional node features into account to guide community discovery and network generation.

\subsubsection{Model} 
In this modeling scenario, we specify HL-MRF priors to utilize feature commonality between nodes to model their community membership. The priors in Table \ref{table:featSimilarityRules} capture that if two nodes have many common features, then there is a greater chance of these nodes being grouped together. The first rule in Table \ref{table:featSimilarityRules} means if node $p$ and node $q$ have the same features (multiple instantiations of Rule 1), then we infer that node $p$ and node $q$ are similar. \emph{similarity(p,q)} is a latent variable that helps us model the degree of similarity between two nodes. The second rule captures that if two nodes are similar, then we can infer that they have similar membership distributions.
\begin{table}%
	\caption{HL-MRF priors to guide community discovery based on presence of multiple common features measured using latent variable \textit{similarity(p, q)}}
	\begin{tabular}{p{8cm}}
	\toprule
	\small{\textbf{Feature-Based Similarity Model}}\\
	\midrule
	\small{feature(p, T) $\land$ feature(q, T) $\land$ link(p, q) $\to$ similarity(p, q)} \\
	\small{similarity(p, q) $\land$ $\pi$(p, K) $\to$ $\pi$(q, K)} \\
	\small{similarity(p, q) $\land$ $\neg \pi$(p, K) $\to$ $\neg \pi$(q, K)} \\
	\bottomrule
	\end{tabular}
    \label{table:featSimilarityRules}
	\vspace{-0.2cm}
\end{table}
\subsubsection{Results} 
We evaluate this model on two Facebook Ego datasets: 1) Ego-414 dataset containing $159$ nodes and $3386$ links, and 2) Facebook Ego-686 dataset containing $170$ nodes and $3312$ links. Table \ref{table:featsim} shows that our model achieves a better training and test log-likelihood and AUC on two Facebook ego datasets when compared with standard MMSB, Copula-MMSB, and IRM.

\begin{figure}[!htb]
\centering
\subfigure[Correlation between similarity value and number of common features]{
\includegraphics[width=0.46\columnwidth]{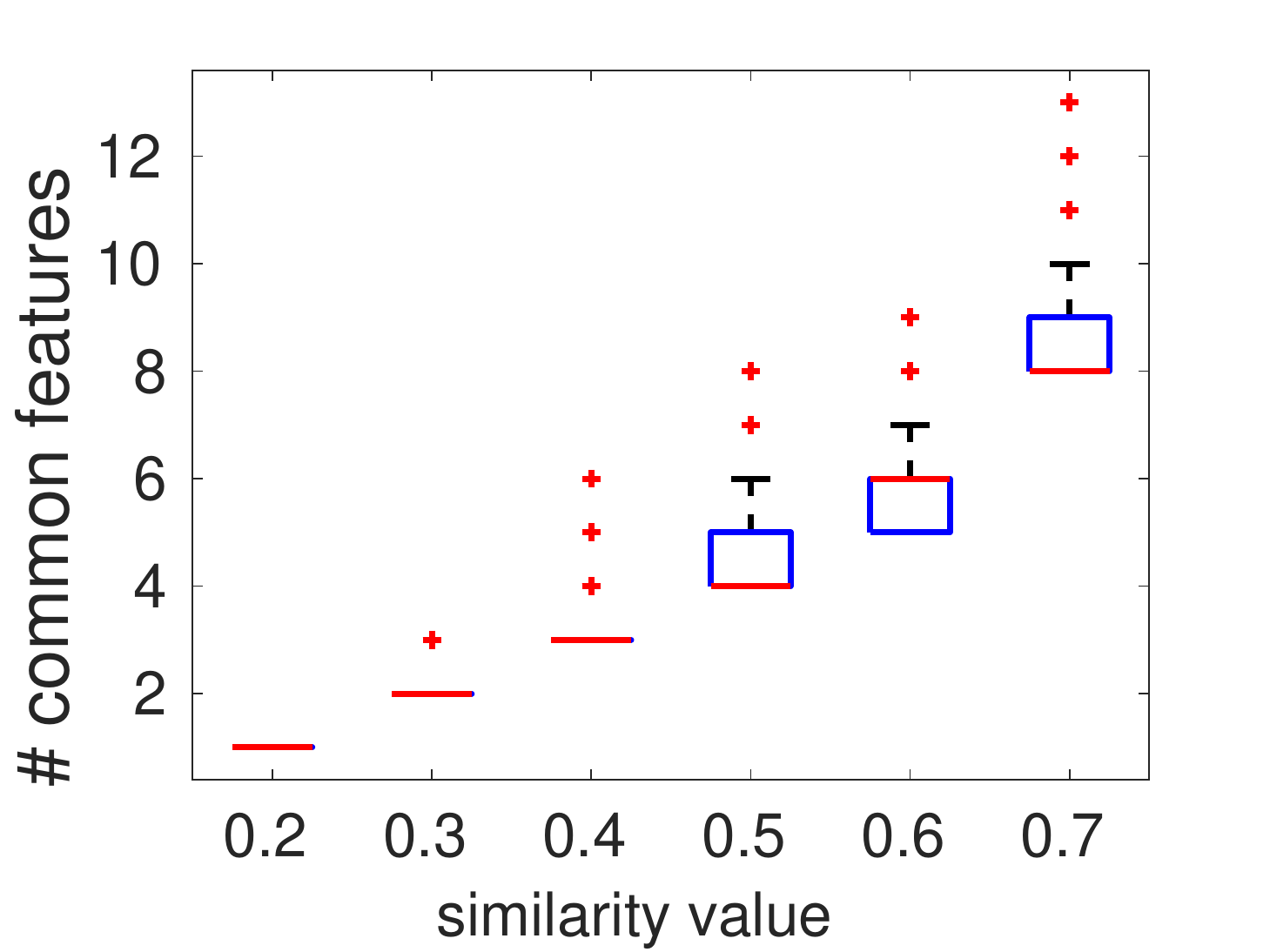}
\label{fig:latentFeat}
}
\centering
\subfigure[Correlation between similarity value and community membership distributions]{
\includegraphics[width=0.46\columnwidth]{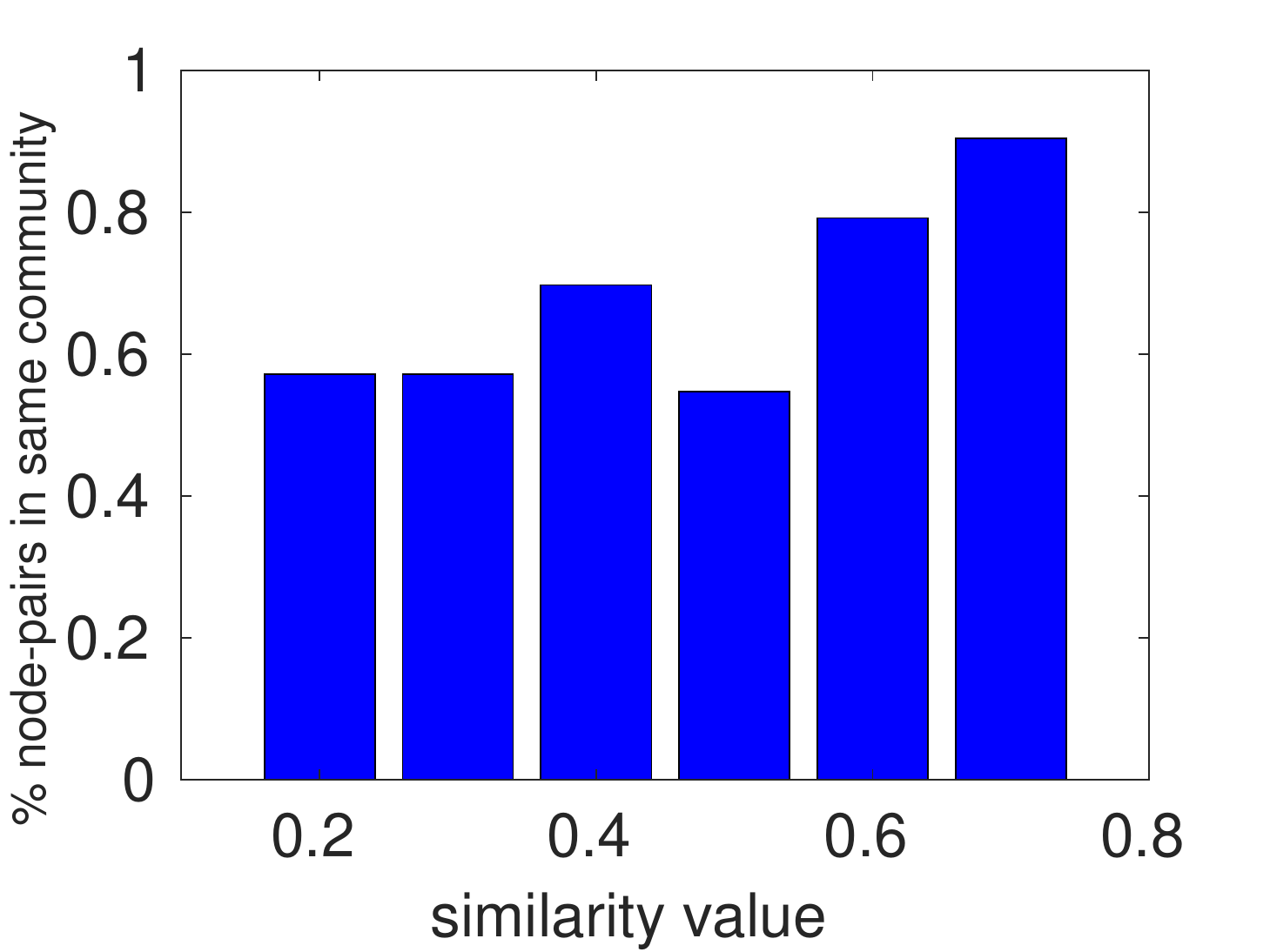}
\label{fig:latentCommunity}
}
\caption{Correlation between latent variable values and membership distributions} 
\label{fig:aucComparison}
\end{figure}

Our latent variables, apart from providing the model with modeling power, also bring interpretability to the model. Figures \ref{fig:latentFeat} and \ref{fig:latentCommunity} illustrate the correlation between latent variable \emph{similarity}  and the number of common features and membership distributions. This conforms with our structured prior that a commonality in features between a pair of nodes can indicate that they belong to the same communities. We also observe that our latent variable values act as a proxy to the relationship between similarity in features and their membership distributions and helps us interpret them better. For example, in Ego-414 dataset, we get the value for the latent variable  \emph{similarity} for nodes $74$ and $88$ to be $0.714$ and we observe that they both have similar membership distributions after training, having a high value for the same community.

\begin{table}
	\caption{Results on Facebook Ego-414 and Ego-686 datasets comparing Struct-MMSB with Copula-MMSB, Standard MMSB, and IRM.}
	\begin{tabular}{p{0.8cm}p{2.2cm}p{1.5cm}p{1.5cm}p{0.5cm}}
	\toprule
	Dataset & Model & Log- & Test Log- & AUC \\
	& & Likelihood & Likelihood & \\
	\midrule
		\multirow{ 3}{*}{Ego414}&\text{Struct-MMSB} & \textbf{-1042.986} & \textbf{-787.406} & \textbf{0.954} \\
	&Copula-MMSB & {-1909.378} & {-1404.075} & {0.844} \\
	&MMSB & {-1309.271} & {-986.321} & {0.9220} \\
	&IRM & {-2205.484} & {-1521.684} & {0.746} \\
	\midrule
	\multirow{ 3}{*}{Ego686}&\text{Struct-MMSB}  & \textbf{-1628.858} & \textbf{-1169.224} & \textbf{0.892} \\
	&Copula-MMSB & {-2002.216} & {-1407.204} & {0.854} \\
	&MMSB & {-1690.593} & {-1221.328} & {0.875} \\
	&IRM & {-2840.591} & {-1866.839} & {0.659} \\
	\bottomrule
	\end{tabular}
    \label{table:featsim}


\end{table}

%
%
\subsection{Case 2: multi-relational graphs} 
In our next experiment, we consider a multi-relational graph setting, where there are multiple different relationships between a single pair of nodes. Here, we experiment on two different kinds of multi-relational data: i) the presence of multiple different types of relationships between a pair of nodes, and ii) the presence of the same type of relationship between a pair of nodes at different timestamps, each timestamp capturing a single relationship.
\subsubsection{Model} \hspace{0.2 cm} To guide the generation of the model in the multi-relational setting, our structured HL-MRF prior captures that if a pair of two nodes $p$ and $q$ have multiple links between them (Rule 1 in Table \ref{table:linkCloseRules}), then, they are ``closer'' than other pairs of nodes. This closeness in their relationship is modeled using the latent variable \emph{close(p, q)}. Rule 2 in  Table \ref{table:linkCloseRules} captures that if two nodes are closer (i.e., have a higher value of latent variable \emph{close(p, q)}), then there is a higher probability of a link existing between them in the network. Note that in Rule 2, we connect the blockmodel $B$ that captures the interaction between communities, membership distributions of nodes $p$ and $q$ with the latent variable \emph{close} induced by the structured prior to guide the generation process.


\begin{table}%
	\begin{centering}
	\caption{Capturing proximity between nodes based on the presence of multiple links between them using latent variable \textit{close(p, q)}}
	\begin{tabular}{l}
	\toprule
	\small{\textbf{Node Proximity in Multi-Relational Graphs}}\\
	\midrule
	\small{link(p, q, T) $\to$ close(p, q)} \\
\small{close(p, q) $\land$ B(K1,K2) $\land$ $\pi$(p, K1) $\to$ $\pi$(q, K2)} \\
	\bottomrule
	\end{tabular}
    \label{table:linkCloseRules}
	\end{centering}
\end{table}
\subsubsection{Results}
\vspace{-0.2cm}
We evaluate the performance of our model on two datasets: 1) Unified Medical Language System dataset (UMLS) consisting of $135$ nodes, $49$ relations, and a total of $6752$ links, and 2) Email-Temporal dataset consisting of $142$ nodes, $525$ timestamps, and a total of $48,141$ links. Table \ref{table:LinkClose} shows the comparison of the performance scores of our model with standard MMSB and Multi-Copula (Copula extended to the multi-relational case to enable a fair comparison) at training and testing time on UMLS and Email-Temporal datasets, where our model achieves better log-likelihood scores on training and test. 

\begin{figure}[!htb]
\centering
\subfigure[Progression toward convergence of batch \text{Struct-MMSB} in UMLS dataset]{
\includegraphics[width=0.46\columnwidth]{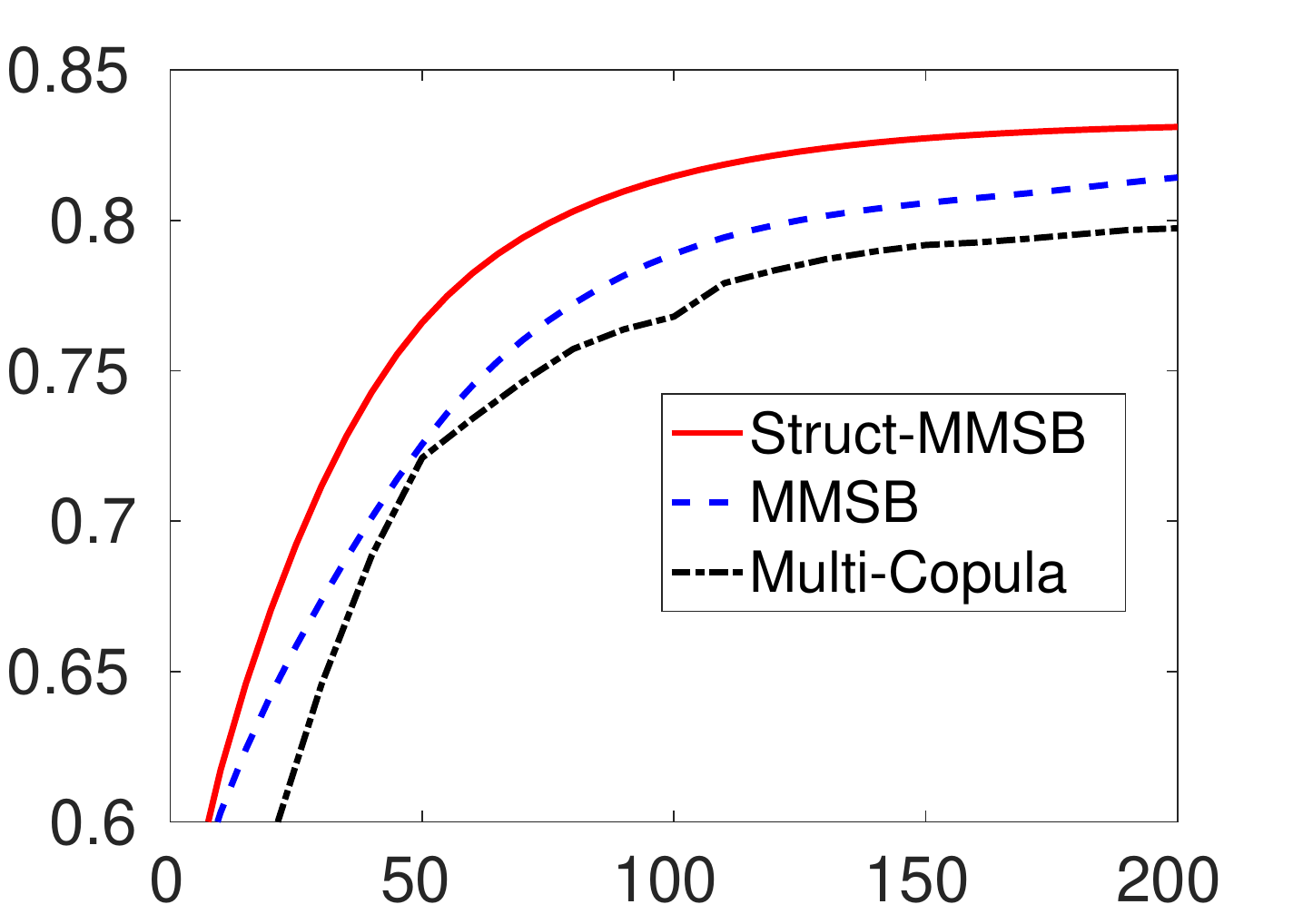}
\label{fig:batch}
}
\subfigure[Progression toward convergence of stochastic \text{Struct-MMSB} in Facebook Ego107]{
\includegraphics[width=0.46\columnwidth]{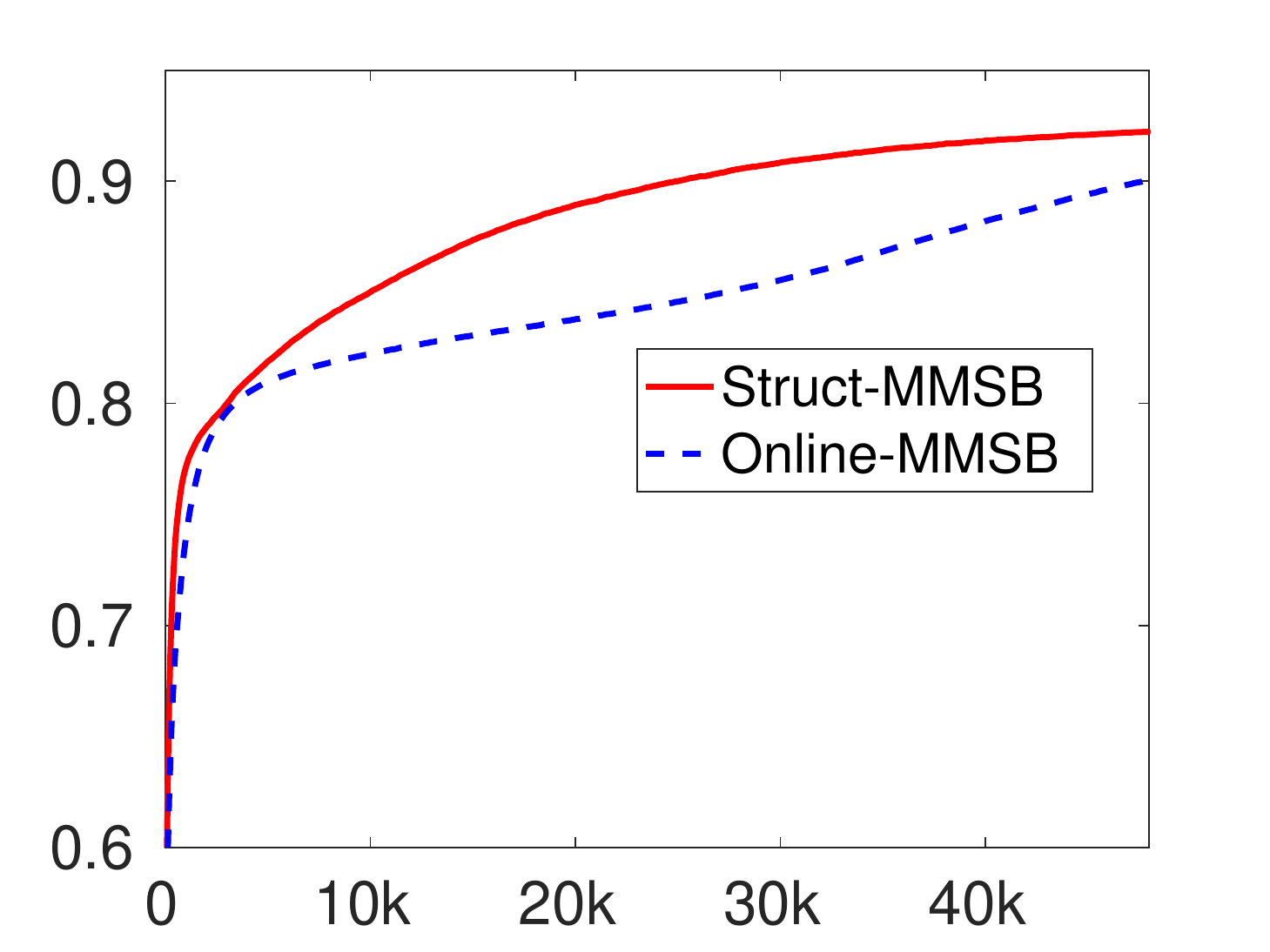}
\label{fig:stochastic}
}
\caption{\text{Struct-MMSB} progresses faster toward convergence in both batch and stochastic scenarios} 
\label{fig:aucComparison}
\end{figure}

Also, as the priors are expected to guide the model in the right direction in the beginning, a better measure of the effectiveness of the structured priors is evaluating the progression toward convergence. Figure \ref{fig:batch} shows the progression of the models toward convergence. We observe that {Struct-MMSB} progresses faster toward convergence and gets a higher AUC value right from the early iterations.

%


\begin{table}
	\begin{centering}
		\caption{Comparing Struct-MMSB with Standard and Multi-Copula MMSB on UMLS and Email-Temporal datasets.}
	\begin{tabular}{p{0.9cm}p{2.3cm}p{1.6cm}p{1.4cm}p{0.5cm}}
	\toprule
	Dataset & Model & Log- & Test Log-& AUC \\
	& & Likelihood & Likelihood & \\
	\midrule
		\multirow{ 3}{*}{UMLS}& \text{Struct-MMSB} & \textbf{-11868.051} & \textbf{-2945.953} & \textbf{0.831} \\
	&Multi-Copula & {-12242.841} & {-3153.354} & {0.785} \\
	&MMSB & -12061.241 & {-2998.037} & 0.814 \\
	\midrule
	\multirow{ 3}{*}{Email}&\text{Struct-MMSB}  & \textbf{-1523.488} & \textbf{-175.871} & \textbf{0.997} \\
		&Multi-Copula & {-1777.112} & {-210.090} & {0.995} \\
	&MMSB & -1562.353 & {-180.620} & 0.995 \\
	\bottomrule
	\end{tabular}
    \label{table:LinkClose}    
	\end{centering}
\end{table}

%



\subsection{Case 3: community discovery using neighborhood similarity}

Here, we consider the problem of discovering communities in graphs using neighborhood similarity. Note that this is the setting for which MMSB was originally conceived. In this experiment, we evaluate the performance of the stochastic Struct-MMSB with the online variant of MMSB on performance and the ability to converge faster.
\subsubsection{Model} The mixed membership $\pi$ is a low rank representation of node's neighborhood. We guide $\pi$ by capturing that if two nodes have multiple common neighbors (Rule $1$ in Table \ref{table:linkSimilarityRules}), then they have similar membership distributions (Rules $2$ and $3$ in Table \ref{table:linkSimilarityRules}). We capture the neighborhood similarity using the latent variable \textit{similarity(p, q)}. 
   \begin{table}%
	\begin{centering}
	\caption{HL-MRF priors capturing neighborhood similarity between nodes based on the presence of multiple neighbors using latent variable \textit{similarity(p, q)}}
	\begin{tabular}{l}
	\toprule
	\small{\textbf{Community Discovery using Neighborhood Similarity}}\\
	\midrule
	\small{link(p, r) $\land$ link(q, r) $\to$ similarity(p, q)} \\
	\small{similarity(p, q) $\land$ $\pi$(p, K) $\to$ $\pi$(q, K)} \\
	\small{similarity(p, q) $\land$ $\neg \pi$(p, K) $\to$ $\neg \pi$(q, K)} \\
	\bottomrule
	\end{tabular}
    \label{table:linkSimilarityRules}
	\end{centering}
\end{table}


\begin{table}
	\begin{centering}
		\caption{Results on Ego-107 and Email-Eu-Core datasets comparing stochastic Struct-MMSB with online MMSB}
	\begin{tabular}{p{0.9cm}p{2.1cm}p{1.5cm}p{1.5cm}p{0.4cm}}
	\toprule
	Dataset & Model & Log- & Test Log- & AUC \\
	& & Likelihood & Likelihood & \\
	\midrule
		\multirow{ 3}{*}{Ego107}&\text{Struct-MMSB} & \textbf{-45202.096} & \textbf{-11441.823} & \textbf{0.922} \\

	&	Online-MMSB & -48534.366 & -12262.921 & 0.900 \\
	\midrule
	\multirow{ 3}{*}{Email}&\text{Struct-MMSB} & \textbf{-31425.953} & \textbf{-8059.303} & \textbf{0.862} \\
	&Online-MMSB & -31562.378 & -8133.698 & 0.861 \\
	\bottomrule
	\end{tabular}
    \label{table:Linksim}
	\end{centering}
\end{table}

\subsubsection{Results} \vspace{-0.2cm} We evaluate performance on the Facebook Ego-107 dataset, which contains $1045$ nodes and $53498$ links and Email-Eu-core dataset, which contains $1005$ nodes and $25571$ links. We implement the stochastic optimization for both \text{Struct-MMSB} and standard MMSB and compare our stochastic \text{Struct-MMSB} with stochastic/online MMSB. For each mini-batch iteration, we randomly sample 10\% of the nodes from all the data. We add all links that exist between the sampled nodes to the mini-batch. In total, we run 48,000 mini-batch iterations.  Our stochastic model achieves better log-likelihood at training and test time when compared to online-MMSB as evident from Table \ref{table:Linksim}. We also observe a similar convergence trend in the stochastic model as well (Figure \ref{fig:stochastic}), where our model progresses faster toward convergence even while only observing a small subset of all the nodes and relationships during each iteration.

\section{Conclusion}
\label{sec:conclusions}

We presented a versatile general-purpose MMSB, \textit{Struct-MMSB}, that is capable of encoding multi-relational data, additional features, and dependencies among the membership distributions, community interaction matrix, and learn meaningful latent variables in the process. We present a batch inference algorithm using EM and a scalable variant using stochastic EM and a method to learn the weights of the structured HL-MRF priors. Our experimental evaluation demonstrates the ability of our model to achieve a superior log-likelihood on training and held-out test data and faster convergence on three different modeling scenarios across 6 datasets and the interpretable nature of our learned latent variables.

%
\bibliographystyle{ecai}


\end{document}